\documentclass[letterpaper, 10 pt, conference]{ieeeconf}  

\IEEEoverridecommandlockouts                              

\usepackage{graphics} 
\usepackage{epsfig} 

\title{\LARGE \bf
Concurrent Tracking of Inliers and Outliers
}

\author{ \parbox{5 in}{\centering Jae-Yeong Lee and Wonpil Yu \\
         Electronics and Telecommunications Research Institute (ETRI)\\
         Daejeon, Korea\\
         {\tt\small \{jylee, ywp\}@etri.re.kr}}
         \hspace*{ 0.5 in}
}

\begin{document}

\maketitle
\thispagestyle{empty}
\pagestyle{empty}

\begin{abstract}

In object tracking, outlier is one of primary factors which degrade performance of image-based tracking algorithms. In this respect, therefore, most of the existing methods simply discard detected outliers and pay little or no attention to employing them as an important source of information for motion estimation. We consider outliers as important as inliers for object tracking and propose a motion estimation algorithm based on concurrent tracking of inliers and outliers. Our tracker makes use of pyramidal implementation of the Lucas-Kanade tracker to estimate motion flows of inliers and outliers and final target motion is estimated robustly based on both of these information. Experimental results from challenging benchmark video sequences confirm enhanced tracking performance, showing highly stable target tracking under severe occlusion compared with state-of-the-art algorithms. The proposed algorithm runs at more than 100 frames per second even without using a hardware accelerator, which makes the proposed method more practical and portable.
\end{abstract}

\section{Introduction}



Object tracking plays a crucial role for successful implementation of various kinds of vision applications such as surveillance, human robot interaction, activity recognition, navigation of intelligent vehicles, and the like. To ensure robust performance, object tracking almost always struggles against dynamic natural scenes.


One of the toughest challenges confronted by any tracking algorithm is to locate a target robustly in the presence of outliers resulting from partial occlusion or background clutter. An outlier can be defined as a statistical observation different in any physical value (for example, color, shape, position, motion vector, and so on) from the others belonging to a target in question. Outliers make it hard not only to locate a target but also to update its object model correctly.


In the object tracking domain, there are roughly two different approaches to deal with outliers. One approach is to use a robust motion estimator\cite{Kalal10a} or a robust similarity measure\cite{Adam06,Shu12}, designed to preserve its original performance to some extent in the presence of outliers. An obvious problem of this approach is that trackers are prone to drift when the ratio of outliers exceeds a estimator-specific threshold. For example, Median Flow tracker \cite{Kalal10a} is based on the assumption that the occluded portion of a target is less than 50 percent of its whole area.


The other approach identifies and removes outliers explicitly to minimize their influence on target localization. An  inherent problem in this kind of approach is how to identify outliers faithfully and how to restore inliers promptly from disocclusion. Previous works \cite{Hariharakrishnan05,Pan07,Amezquita08} mainly focus on the detection of occlusion and disocclusion, showing reactive behavior of target tracking depending on instantaneous scene structure.

\begin{figure}[t]
\begin{center}
   \includegraphics[width=1.0\linewidth]{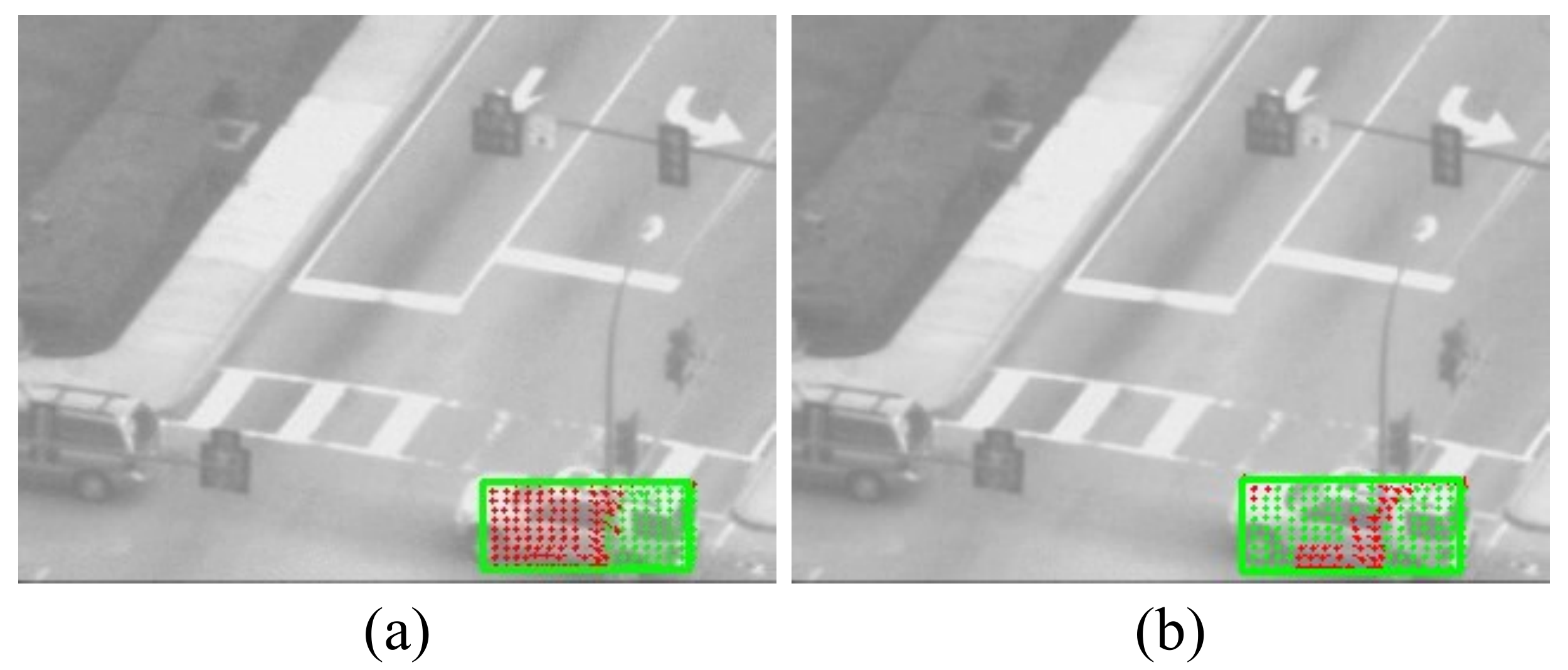}
\end{center}
   \caption{(a) The result of inlier restoration by using only motion difference. (b) The result of inlier restoration by using both motion difference and reference appearance model. Tracking points whose state is inlier are depicted by green color and outlier by red color.}
\label{fig:restoration_problem}
\end{figure}

In this paper we consider outliers as important as inliers for object tracking and propose a motion estimation algorithm based on concurrent tracking of inliers and outliers. Our approach is quite different from most of the existing methods which simply discard detected outliers and pay little or no attention to employing them as an important source of information for motion estimation.

The rest of the paper is organized as follows. In Section \ref{sec:concurrent_tracker}, we present our approach and give a detailed description of its individual parts. In Section \ref{sec:experiment}, we evaluate the performance of our proposed tracker and compare the result to other state-of-the-art methods on benchmark tracking data sets. Finally, in Section \ref{sec:conclusion}, we conclude the paper.

\section{Concurrent tracker}
\label{sec:concurrent_tracker}

\begin{figure}[t]
\begin{center}
   \includegraphics[width=1.0\linewidth]{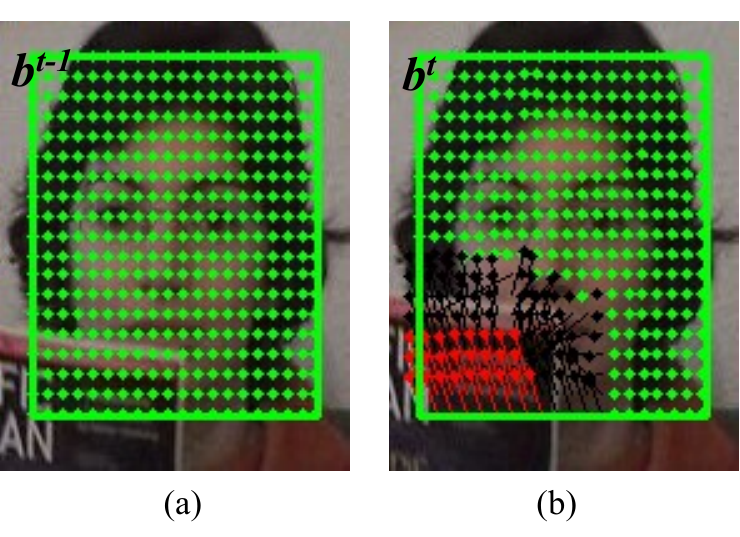}
\end{center}
   \caption{Estimation of optical flow and estimation of robust transformation.}
\label{fig:basic_tracker}
\end{figure}

\subsection{Basic flow tracker}
\label{ssec:basic_tracker}

Our tracker is similar to Median-Flow tracker \cite{Kalal10a} in estimating target motion based on optical flow and making use of pyramidal implementation of the Lucas-Kanade tracker \cite{Bouguet99}.


The proposed tracker accepts as input a pair of images $I^{t-1}$, $I^t$ and a bounding box $b^{t-1}$ surrounding a target in question, either determined manually or determined automatically by a target detection algorithm prior to target tracking. The proposed tracker outputs an updated estimate of the target location in the form of a bounding box $b^t$. A set of grid points $\{p_i^{t-1}\}$ is initialized uniformly across the bounding box $b^{t-1}$ (Figure \ref{fig:basic_tracker}a). For these points, optical flow is computed by Lucas-Kanade tracker. Let $\{(p_i^{t-1}, \hat{p}_i^{t})\}$ be a set of estimated flows. Unreliable flows are then identified based on the Forward-Backward (FB) error criterion introduced in \cite{Kalal10a}.

For evaluation of FB error, the proposed tracker computes the optical flow backwardly from $I^t$ to $I^{t-1}$ for the set of tracked points $\{\hat{p}_i^{t}\}$, which generates a set of backward flows $\{(\hat{p}_i^{t}, \hat{p}_i^{t-1}*)\}$. The FB error is computed by employing Euclidean distance between  ${p}_i^{t-1}$ and $\hat{p}_i^{t-1}*$ and the flows whose FB error is larger than a predefined threshold $\delta$ are considered as unreliable and eventually filtered out. We will call the grid points $p_i^{t-1}$ associated with the remaining reliable flow {\it matched} points and the others {\it unmatched} points. Note that the tracked points may fail to be matched during optical flow estimation by the Lucas-Kanade tracker. We do not distinguish such unmatched points from the points filtered out and simply call both of them {\it unmatched} points.

Considering only in-plane rotation, scale change, and translation, a rigid transformation $T$ is then estimated from the matched motion flows by RANSAC \cite{Fischler81} estimator. We adopt rigid transformation because more general transformation such as full affine transformation or homography is more error-prone for most tracking tasks. Finally the bounding box $b^t$ is obtained by applying the estimated transformation to $b^{t-1}$:
\begin{equation}
b^t = T(b^{t-1}).
\label{eq:box_update}
\end{equation}

Figure \ref{fig:basic_tracker} illustrates the tracking procedure of the proposed tracker. In Figure \ref{fig:basic_tracker}b, unreliable flows which are filtered out by FB test with a threshold value $\delta = 1.5$ is depicted in black while the matched flows are depicted in green or red according to whether it is classified as inlier (green) or outlier (red) by RANSAC estimator.

\subsection{Concurrent tracking of inliers and outliers}

The basic flow tracker described in Section \ref{ssec:basic_tracker} has a limited performance for several reasons although it is robust to outliers to some extent. For example, the tracker may fail even in the presence of a small fraction of outliers if the outliers have motion flows similar to those of the inliers and not filtered out since a small drift can lead a tracker to fail for long-term operation. This situation is not rare in actual tracking tasks.

In addition, the tracker may fail if the number of outliers exceeds the number of inliers and the outliers form consistent motion flows as in the case of partial occlusion. In such case the tracker will drift along a direction of the outlier motion. A more severe problem is that the tracker may not recognize a tracking failure when target motion estimation is carried out in a normal manner.

We solve the aforementioned problems by extend the functionality of the basic flow tracker to track inliers and outliers separately at the same time, which we call the basic flow tracker with the extended functionality a {\it concurrent tracker} (or simply a tracker in the remaining part of this paper). Let $p_i^{t-1}$ be tracking points and $s_i^{t-1} = s(p_i^{t-1})$ be the states of the tracking points at time $t-1$, where
\begin{equation}
s(p) =  \bigg\{ \begin{array}{cc}
		 1, &  p = inlier\\
		 0, & p = outlier
	 \end{array}.
\label{eq:state}
\end{equation}

Initially, the tracking points $p_i^0$ are initialized by uniform grid points across the bounding box $b^0$ and their states are set to be {\it inlier} ($s_i^0 = 1$). The concurrent tracker accepts a pair of images $I^{t-1}$, $I^t$ and a bounding box $b^{t-1}$ and a model state $S^{t-1}$ and outputs $b^t$ and $S^t$, where $S^t = \{s_i^t\}$. The tracking points $p_i^{t-1}$ always are reinitialized by the uniform grid points within the bounding box $b^{t-1}$. The tracker firstly estimates optical flows and then filters out unreliable flows as in the basic flow tracker of section \ref{ssec:basic_tracker}.

Let $P_{unmatch} = \{p_j^{t-1}\}$ be the set of unmatched points; $P_{match} = \{p_i^{t-1}\}$ be the set of  matched points; and $F_{match} = \{(p_i^{t-1}, \hat{p}_i^{t})\}$ be the set of matched flows. The set of matched points  $P_{match}$ are further divided into two sets of matched inliers $P_{in} = \{p_i^{t-1}|s_i^{t-1} = 1,~~ p_i^{t-1} \in P_{match}\}$ and matched outliers $P_{out} = \{p_i^{t-1}|s_i^{t-1} = 0,~~ p_i^{t-1} \in P_{match}\}$. Similarly $F_{match}$ is divided into $F_{in}$ and $F_{out}$.

Next, the tracker estimates two rigid transformations of $T_{in}$ for inlier flows $F_{in}$ and $T_{out}$ for outlier flows $F_{out}$ by applying RANSAC estimator respectively to the two sets of matched flows. If $T_{in}$ is estimated successfully ({\i i.e.,} in case the number of matched inlier flows is sufficiently large for estimating the rigid transformation and the estimated transformation is supported by a number of matched inlier flows more than a predefined threshold), the tracker updates the bounding box as follows:
\begin{equation}
b^t = T_{in}(b^{t-1}).
\label{eq:box_update_success}
\end{equation}
If it fails to estimate $T_{in}$, the tracker fails ({\i i.e.,} in the event of the insufficient number of matched flows for estimating the rigid transformation or the estimated transformation is supported by too small number of inliers) and stops all the remaining steps, and simply returns the previous state:
\begin{equation}
b^t = b^{t-1},
\label{eq:box_update_fail}
\end{equation}
\begin{equation}
S^t = S^{t-1}.
\label{eq:state_update_fail}
\end{equation}

In case we succeed to estimate both of $T_{in}$ and $T_{out}$, the states of tracking points are updated as follows. First, those flows far away from the estimated transformation are determined as unreliable and filtered out. Let $P_{in}^*$ and $P_{out}^*$ be the remaining flow points after the filtering, given by
\begin{equation}
P_{in}^* = \{ p_i^{t-1} | p_i^{t-1}\in P_{in}, r((p_i^{t-1},\hat{p}_i^t),T_{in}) < k_{in}\sigma_{in}\}
\label{eq:residual_reject_in}
\end{equation}
\begin{equation}
P_{out}^* = \{ p_i^{t-1} | p_i^{t-1}\in P_{out}, r((p_i^{t-1},\hat{p}_i^t),T_{out}) < k_{out}\sigma_{out}\},
\label{eq:residual_reject_out}
\end{equation}
where $r$ is residual given by
\begin{equation}
r((p^{t-1},\hat{p}^t),T) = |T(p^{t-1}) - \hat{p}^t|
\label{eq:residual}
\end{equation}
and $\sigma_{in}^2$ and $\sigma_{out}^2$ are the estimated variance of the inlier residuals and outlier residuals respectively, and $k_{in}$ and $k_{in}$ are constants.

The points in $P_{in}^*$ and $P_{out}^*$ can be regarded as error-free and the model state is updated based on these points.
The model state is updated according to the following rules:
\begin{enumerate}
\item $s(p)$ is set to be {\it outlier} if $p \in P_{out}^*$.
\item $s(p)$ is set to be {\it inlier} if $p \in P_{in}^*$ and $p \notin P_{out}^*$.
\item $s(p)$ is set to be {\it outlier}, otherwise.
\end{enumerate}

In actual implementation, before the state update, the inlier points in $P_{in}^*$ are smoothed by median filtering to remove noise and then dilated. The median filtering and dilation is applied to a rectangular grid after mapping the points on the rectangular grids, which gives a smoothed version of $P_{in}^*$. The outlier points in $P_{out}^*$ are also smoothed by median filtering but not dilated. After applying smoothing, some inlier points in $P_{in}^*$ and outlier points in $P_{out}^*$ may overlap; in this case, outlier points always have a higher priority according to the update rules above.

In case it fails to estimate $T_{out}$, the model state is updated according to the following rules:
\begin{enumerate}
\item $s(p)$ is set to be {\it inlier} if $p \in P_{in}^*$.
\item $s(p)$ is set to be {\it outlier}, otherwise.
\end{enumerate}

\subsection{Restoration of inliers from motion difference}
\label{ssec:restore_motion}

\begin{figure}[t]
\begin{center}
   \includegraphics[width=1.0\linewidth]{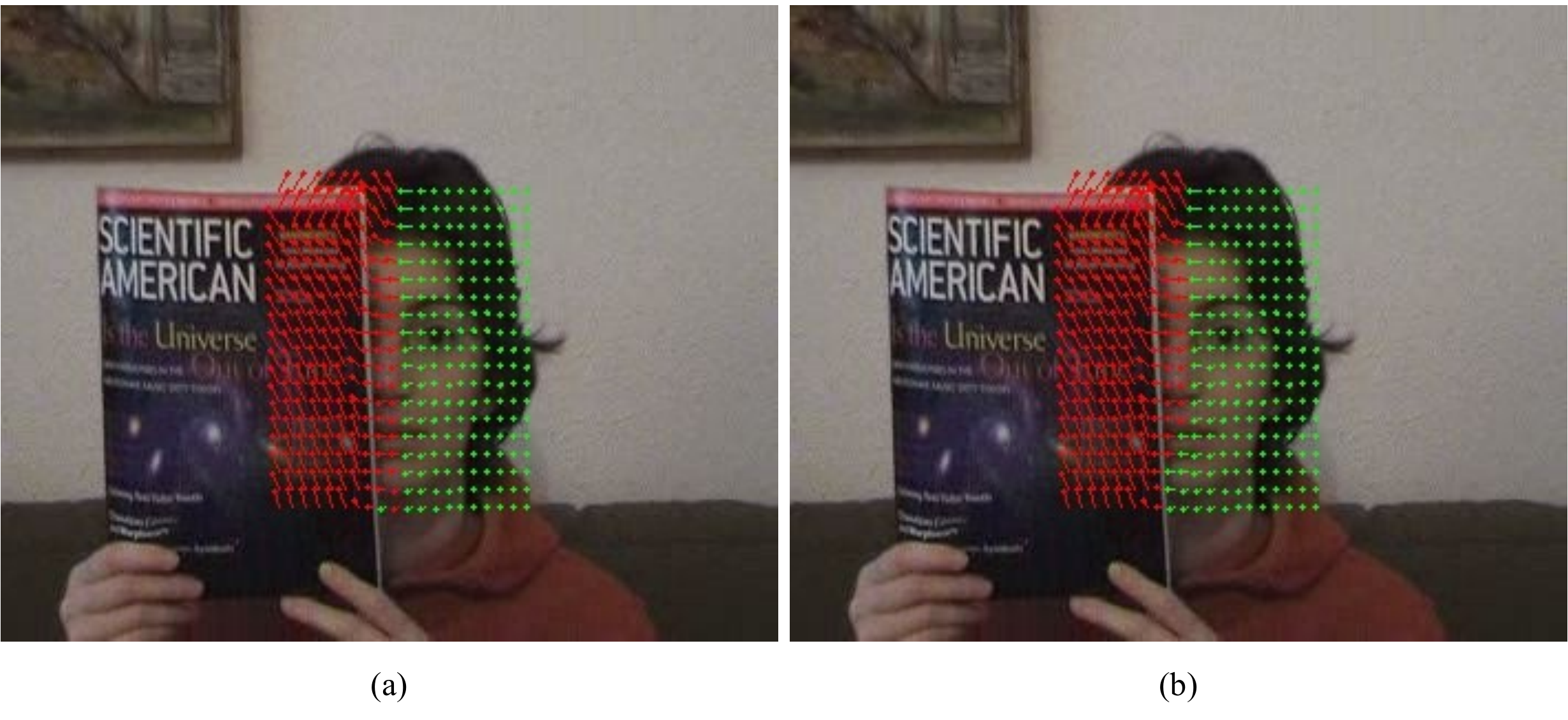}
\end{center}
   \caption{Example of restoration from motion difference. (a) State of grid points before motion restoration. (b) State of grid points after motion restoration.}
\label{fig:motion_restoration}
\end{figure}

Concurrent observation of inlier motion and outlier motion provides a powerful cue for the restoration of target inliers from disocclusion. In case inlier motion and outlier motion are successfully estimated and the two estimated motions are sufficiently different from each other, one of the following two cases holds. One case is that the target is currently occluded by an occluding object and the target and the occluding object have different motion. The other case is that current target ROI (Region Of Interest) includes part of its background and the target is moving while the target ROI contains part of the background. In either case, if a large motion difference is detected, restoration conditions are checked to determine the current outliers as inliers. Although the primary goal of this step is to restore disoccluded target region as inliers, the inliers currently misclassified as outliers due to classification errors can also be restored through the step.

Although a motion difference between inliers and outliers is detected from the step above, if the difference is not sufficiently large, there is a risk of wrong restoration which leads to drift or tracking failure. For example, if outliers have a large motion variance, some outliers can be classified as inliers. Therefore, it is very important to measure motion difference reliably. A simple checking based on Euclidean distance between two average motion vectors of inliers and outliers is not adequate. It is because pure rotation or scale change without translation gives zero mean motion and Euclidean distance does not reflect motion variance. Therefore, in order to estimate a statically meaningful motion difference, we measure distance of two rigid transformations instead of comparing motion vectors directly. The distance is measured based on Mahalanobis distance under the assumption that the motion vectors follow Gaussian distribution.

Mahalanobis distance between inlier motion and outlier motion is estimated as follows:
\begin{equation}
\lambda = {{d^*(T_{in}, T_{out})} \over {\sigma^*}}
\label{eq:mahal_d}
\end{equation}
\begin{equation}
d^*(T_{in}, T_{out}) = {{1 \over n} \sum_{i=1}^n{|T_{in}(p_i^{t-1})-T_{out}(p_i^{t-1})|}}
\label{eq:mahal_d}
\end{equation}
\begin{equation}
\sigma^{2*} = {{n_{in} \sigma_{in}^2 + n_{out} \sigma_{out}^2} \over {n_{in} + n_{out} - 2}},
\label{eq:common_sigma}
\end{equation}
where $d^*$ is the estimated distance between two transformations, $T_{in}$ is the estimated transformation from the matched inlier motions between the previous frame and current frame and  $T_{out}$ is the estimated transformation from the matched outlier motions respectively. $p_i^{t-1}$ is the location of the tracking points in the previous frame and $\sigma^{2*}$ is the estimated common variance of the two distributions of inliers and outlier motions. In (\ref{eq:common_sigma}), $\sigma_{in}^2$ ($\sigma_{out}^2$) is the variance of the residuals of the matched inliers (outliers) with respect to $T_{in}$ ($T_{out}$) and it is computed by

\begin{equation}
\sigma_{in}^2 = {1 \over n_{in}} \sum_{p_i^{t-1} \in P_{in}}{|T_{in}(p_i^{t-1}) - \hat{p}_i^t |^2},
\label{eq:sigma}
\end{equation}
where $n_{in} = |P_{in}|$ is the number of matched inliers.

The condition for inlier restoration is satisfied when the estimated Mahalanobis distance $\lambda$, estimated distance of the two transformations $d^*(T_{in}, T_{out})$, and the ratio of outliers are larger than predefined thresholds respectively:

\begin{equation}
C = (\lambda>{\lambda}_{\theta}) \land (d^*>d_\theta) \land \left({n_{out} \over N} > \alpha \right).
\label{eq:condition}
\end{equation}

In case the condition $C$ is satisfied, we classify the matched outliers whose overall motion is more close to $T_{in}$ than $T_{out}$ and its residual from $T_{in}$ is less than a predefined threshold as inliers.  Finally, state of tracking points is updated as follows:

\begin{equation}
S^t(p_i) =  \bigg\{ \begin{array}{cc}
		 inlier, &  C \land ({\lambda_{in} < \lambda_{out}}) \land ({\lambda_{in}< k_{in}})\\
		 S^{t-1}(p_i), & otherwise
	 \end{array},
\label{eq:restore}
\end{equation}
where
\begin{equation}
\lambda_{in} = \frac{|T_{in}(p_i^{t-1}) - \hat{p}_i^t |}{\sigma_{in}},
\label{eq:residual_in}
\end{equation}
\begin{equation}
\lambda_{out} = \frac{|T_{out}(p_i^{t-1}) - \hat{p}_i^t |}{\sigma_{out}}.
\label{eq:residual_out}
\end{equation}

In our implementation, the parameters are empirically set to be $\lambda_\theta = 3$, $d_\theta = 1.5$, $\alpha = 0.3$, and $k_{in} = 3$.

Figure \ref{fig:motion_restoration} shows an example of inlier restoration based on motion cue. The left figure shows the state of inliers and outliers before applying restoration and the right figure shows an updated state after restoration. In this example, estimated measures were $\lambda = 36.5$, $d^* = 9.6$, and 12 inliers were restored.

\subsection{Restoration of inliers from appearance reference model}

One problem of the restoration precedure using motion difference described in subsection \ref{ssec:restore_motion} is that the restoration is performed only when a significant motion difference occurs. This restoration condition can cause problems in two respects.

Firstly, inliers may be misclassified as outliers by various causes such as abrupt illumination change, 3D motion, and deformation of target as well as occlusion. In that case the restoration condition cannot be met if there is no external occluding object or included background.

Secondly, as the restoration is delayed until the ratio of outliers exceeds some threshold by (\ref{eq:condition}), the rate of unrecovered inliers possibly exceeds true outliers when the size of an occluding object is relatively small as illustrated in Figure \ref{fig:restoration_problem}, which leads to wrong estimation of outlier motion, $T_{out}$. In Figure \ref{fig:restoration_problem}, a moving car crosses the road and the tracking points occluded by a street lamp turn into outliers rapidly. In this case, however, disoccluded tracking points cannot be recovered as the majority of outliers consists of misclassified inliers and thus they will give a motion estimate similar to inlier motion.

To overcome these difficulties, as a complementary step, we perform an additional step of inlier restoration by using an appearance reference model after the restoration step based on motion difference. The restoration procedure using a reference model is as follows.

The appearance reference model is initialized by the initial target region of the image and the state of grid points of the reference model all set to be inliers. We first compute optical flow between the reference model and the estimated target region in a current frame for the grid points of the reference model. Next, we check whether the grid points of the reference model succeeds to match or not and set the state of current tracking point to be inlier if the match succeeds and the state of matched grid point of the reference model is inlier.
\subsection{Motion drift}

Frame to frame motion estimation based on optical flow is prone to drift in the presence of background interference, occlusion, fast illumination changes, and deformation of a target. Even a minute drift at early stage can lead a tracker to totally fail in long term tracking. The appearance reference model described earlier is used not only to restore inliers but also to compensate for possible motion drift of the tracker.

Let $\{(p_i, p_i')\}$ be the matched point pairs between the reference model and the current frame in the previous inlier restoration step, where $p_i$ is the inlier points in the reference model whose state is inlier and $p_i'$ is the matched point in the current frame from the estimation of optical flow. If there exists a rigid transformation and the number of matched points is larger than a predefined threshold, the motion drift is compensated by updating current target location to be the transformed location of the reference location by the estimated transformation.

A subtle and well-known problem caused from the use of a reference model is how to update the model to adapt to the appearance change. Fixing the reference model with the initial model does not suffer from drift or occlusion but cannot adapt to the change of the target. On the other hand, too frequent update of the model may lose its basic role as a reference. Our strategy is that the reference model is updated only when the mean intensity or scale of the current target estimate is different from the reference model significantly more than 10 percent and the ratio of outliers is less than a predefined threshold, $\gamma = 0.2$. The actual update of the reference model is done simply by replacing it with the current target region and the state of the tracking points.

The idea of compensation of motion drift by using an appearance reference model is motivated from the work by Zhang {\i et.al.,} \cite{Zhang12}, where a KLT feature tracker is combinatorially used to detect occlusion and remedy drift of color-based particle filter tracker.

\section{Experiment}
\label{sec:experiment}

\begin{table}
\begin{center}
\begin{tabular}{|c|c|}
\hline
Sequence & Main Challenges\\
\hline\hline
ft\_face \cite{Adam06} & large occlusion for a long time\\
lot\_david \cite{Oron12} & varying illumination, 3D motion\\
lot\_face \cite{Oron12} & occlusion\\
lot\_girl \cite{Oron12} & 3D rotation\\
lot\_human \cite{Oron12} & moving target with full occlusion\\
lot\_shirt \cite{Oron12} & large deformation\\
lot\_shop \cite{Oron12} & partial occlusion\\
lot\_sylv \cite{Oron12} & 3D motion\\
mil\_coke11 \cite{Babenko09} & occlusion, 3D rotation\\
mil\_dollar \cite{Babenko09}& similar background\\
oal\_dudek \cite{Ross08} & temporal full occlusion\\
prost\_board \cite{Santner10} & large inclusion of background\\
prost\_box \cite{Santner10} & occlusion\\
tld\_car \cite{Kalal10b} & moving target with full occlusion\\
tld\_volkswagen \cite{Kalal10b} & long disappear from camera\\
\hline
\end{tabular}
\end{center}
\caption{Test dataset.}
\label{table:dataset}
\end{table}

In this section, we evaluate the performance of the proposed tracker on publicly available dataset. We also compare our tracker to current state of the art trackers. The parameters are not tuned and always the same parameters are used throughout the experiment. All tests were performed on a Intel i7 940 2.93GHz 3GB RAM desktop without aid of GPU acceleration.

\subsection{Implementation}
We use a grid of $m \times m$ tracking points where $m$ is determined adaptively from 10 to 20 according to the target size. The optical flow for them is computed by using a pyramidal implementation of the Lucas-Kanade tracker \cite{Bouguet99}. We set Lucas-Kanade tracker to use two levels of the pyramid and represent the points by $w \times w$ patches, where $w$ is set to be one third of the initial target size. 

\subsection{Quality measure}
We compare the performance of tracking algorithm by using two quality measures of tracking accuracy and success rate. We measure the tracking accuracy by using the detection-criterion of the VOC challenge \cite{Everingham09} given by
\begin{equation}
\lambda_{acc} = {{area(b_d \cap b_{gt})} \over {area(b_d \cup b_{gt})}},
\label{eq:accuracy_measure}
\end{equation}
where $b_d$ denotes a detected bounding box and $b_{gt}$ the ground truth bounding box.

The second measure of success rate is computed by a ratio of the correctly tracked frames. A trajectory bounding box is considered correct if it overlaps with ground truth larger than 50\% ({\i i.e.,} $\lambda_{acc} > 0.5$).

\subsection{Test dataset}
Throughout the experiments, we use publicly available fifteen benchmark video sequences (Table \ref{table:dataset}). We denote the original source of the test video sequences from which the video and ground truth data are downloaded as the prefix of the sequence name. The test dataset includes various tracking challenges such as large occlussion for a long time (ft\_face), full occlusion ('oal\_dudek'), 3D motion or self-occlusion (lot\_girl), temporal disappearance from the camera view ('tld\_car, tld\_volkswagen'), background interference ('prost\_board'), deformation ('lot\_shirt), illumination ('lot\_david') and scale change. Note that the most common challenge of the test video sequences which we focuse on is the occlusion.

\subsection{Performance of the proposed tracker}

\subsubsection{Qualitative performance}

\begin{table}
\begin{center}
\resizebox{1.0\linewidth}{!}{\includegraphics{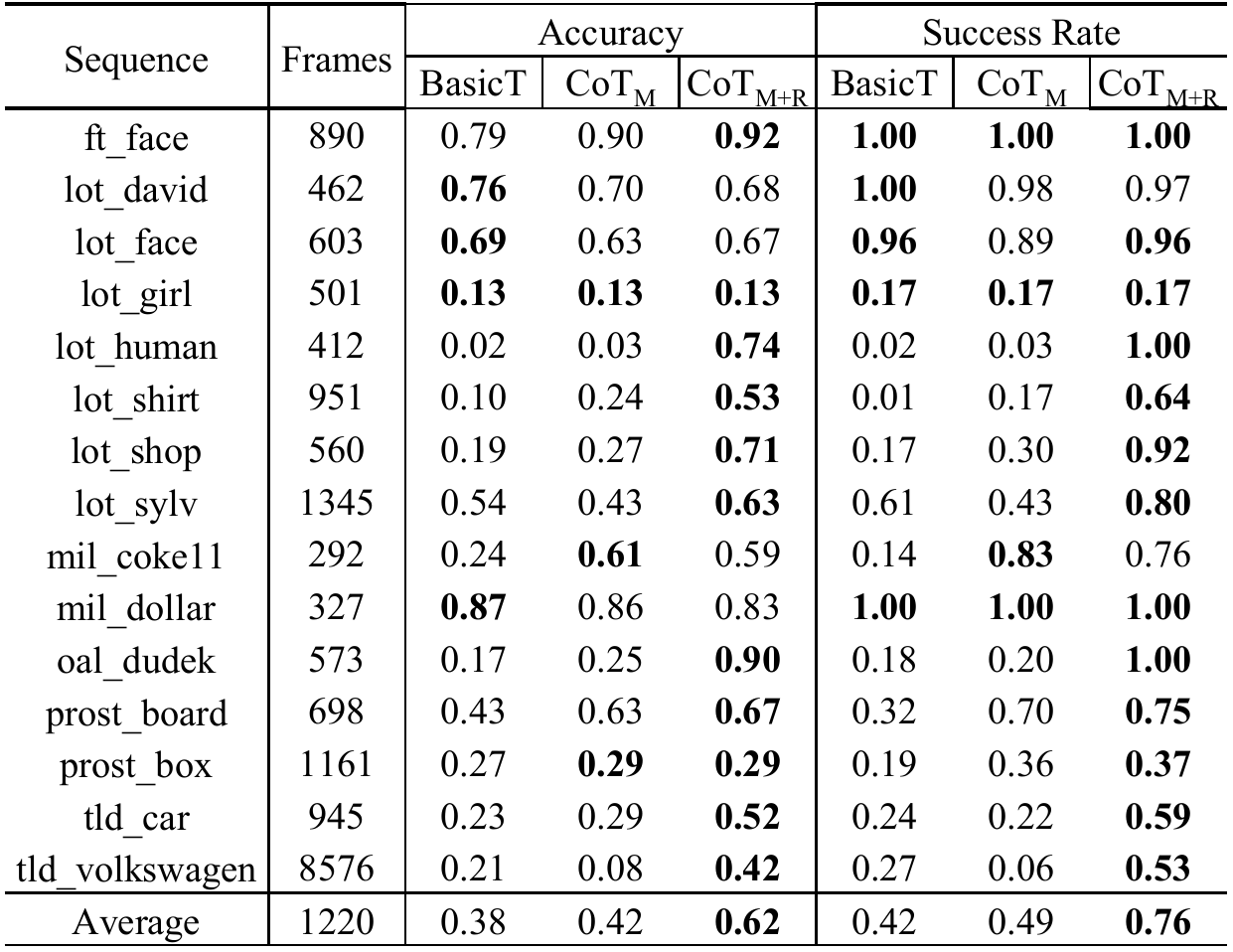}}
\end{center}
\caption{Performance of the proposed tracker.}
\label{table:performance_lee}
\end{table}

In the first experiment, we evaluate the proposed tracking algorithm in term of tracking accuracy and the ratio of correctly tracked frames with three different versions of trackers: basic flow tracker (BasicT), concurrent tracker with motion restoration feature ($CoT_M$), concurrent tracker with motion restoration and appearance reference model ($CoT_{M+R}$). The primary purpose of the first experiment is to investigate the influence of the motion restoration and appearance reference model on the tracking performance. Table \ref{table:performance_lee} shows the experimental results.

We are able to observe that by using concurrent version of trackers the tracking performance overall increases significantly compared to basic flow tracker and gives the best performance when all features (motion restoration and appearance reference model) are enabled.

\begin{figure*}[t]
\begin{center}
   \includegraphics[width=1.0\linewidth]{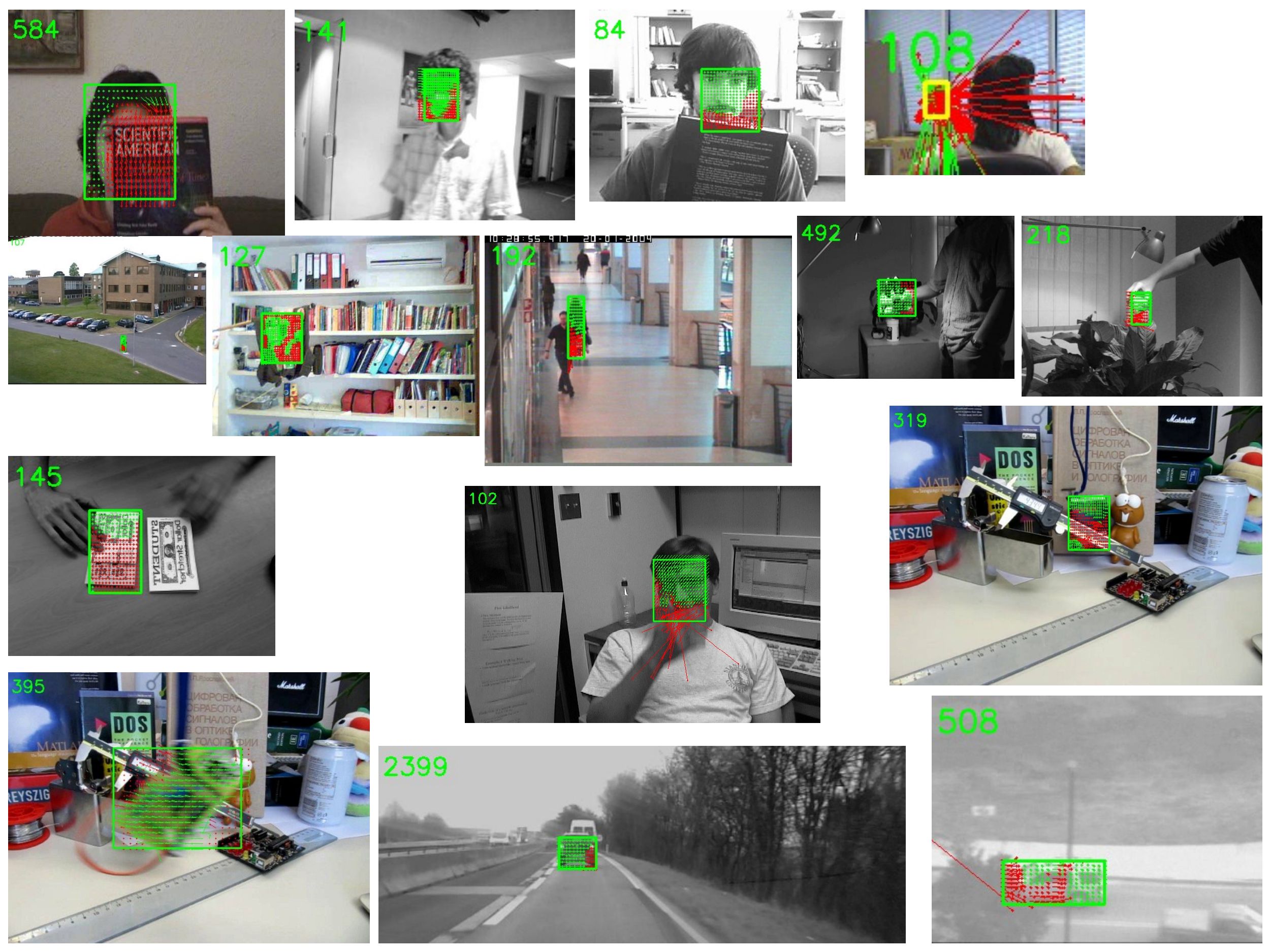}
\end{center}
   \caption{Sample snapshots of our concurrent tracker on the test dataset. The frame number is depicted on the snapshot. Sequence name: (first row) ft\_face, lot\_david, lot\_face, lot\_girl, (second row) lot\_human, lot\_shirt, lot\_shop, lot\_sylv, mil\_coke11, (third row) mil\_dollar, oal\_dudek, prost\_box, (fourth row) prost\_board, tld\_volkswagen, tld\_car.}
\label{fig:snapshot}
\end{figure*}

Figure \ref{fig:snapshot} shows sample snapshots of our concurrent tracker on the test dataset. We are able to observe that the proposed tracker locates the tracking targets very accurately without any drift for most challenging cases except 'lot\_girl' case where it fails due to full self-occlusion of the target.

\subsubsection{Runtime performance}

\begin{table}
\begin{center}
\begin{tabular}{|c|c|c|c|}
\hline
Method & BasicT & $CoT_M$ & $CoT_{M+R}$ \\
\hline\hline
Runtime (ms) & 5 & 6.1 & 7.4 \\
Fps &  200 & 164 & 135 \\
\hline
\end{tabular}
\end{center}
\caption{Average runtime of the proposed tracker on test dataset.}
\label{table:runtime}
\end{table}

Table \ref{table:runtime} shows the average runtime of the proposed trackers on the test dataset. We are able to observe that even with $CoT_{M+R}$ tracker it runs at more than 100 frames per seconds without using any hardware acceleration, which is rarely observable in other trackers. This real time performance makes the proposed method more attractive and portable.

\subsection{Comparison with state of the art trackers }

\begin{table}
\begin{center}
\resizebox{1.0\linewidth}{!}{\includegraphics{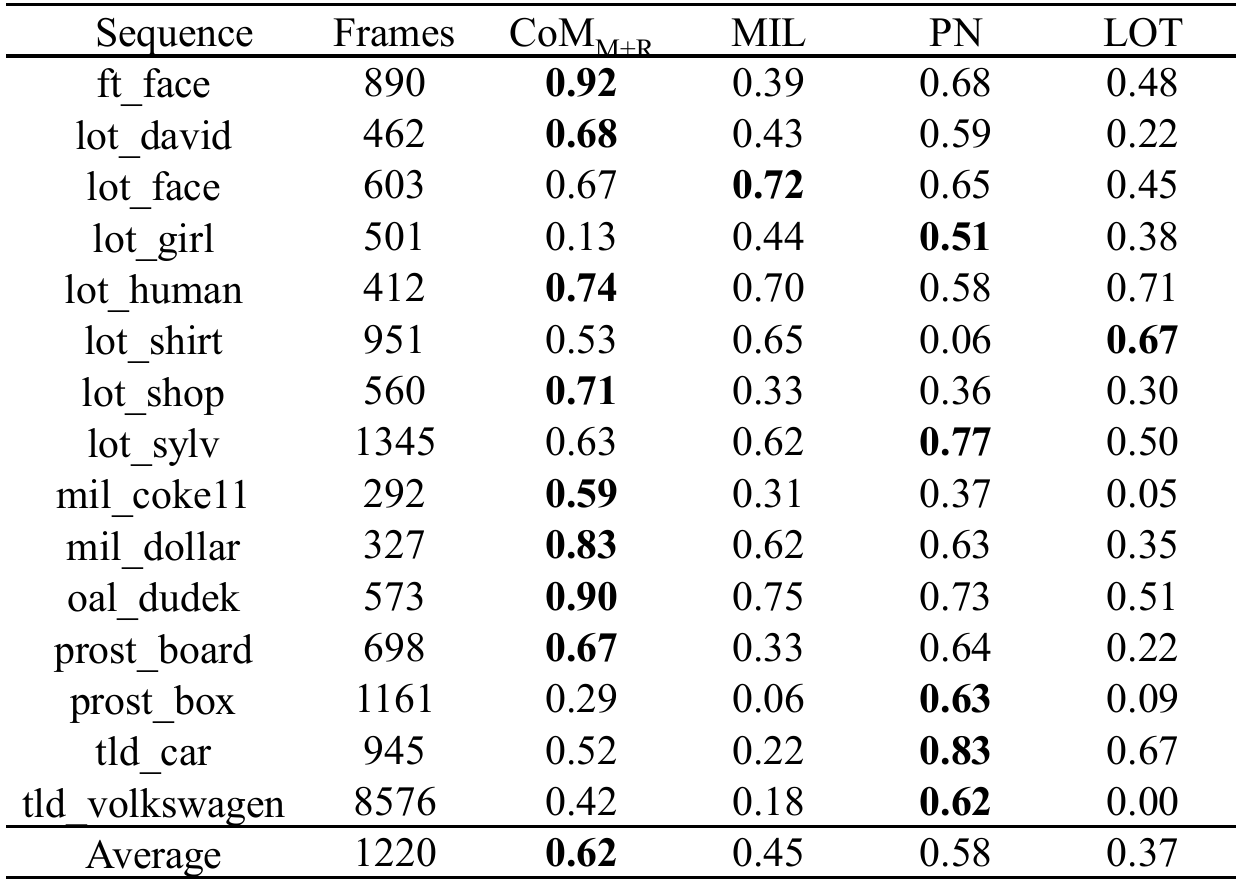}}
\end{center}
\caption{Comparison of tracking accuracy.}
\label{table:accuracy_comp}
\end{table}

\begin{table}
\begin{center}
\resizebox{1.0\linewidth}{!}{\includegraphics{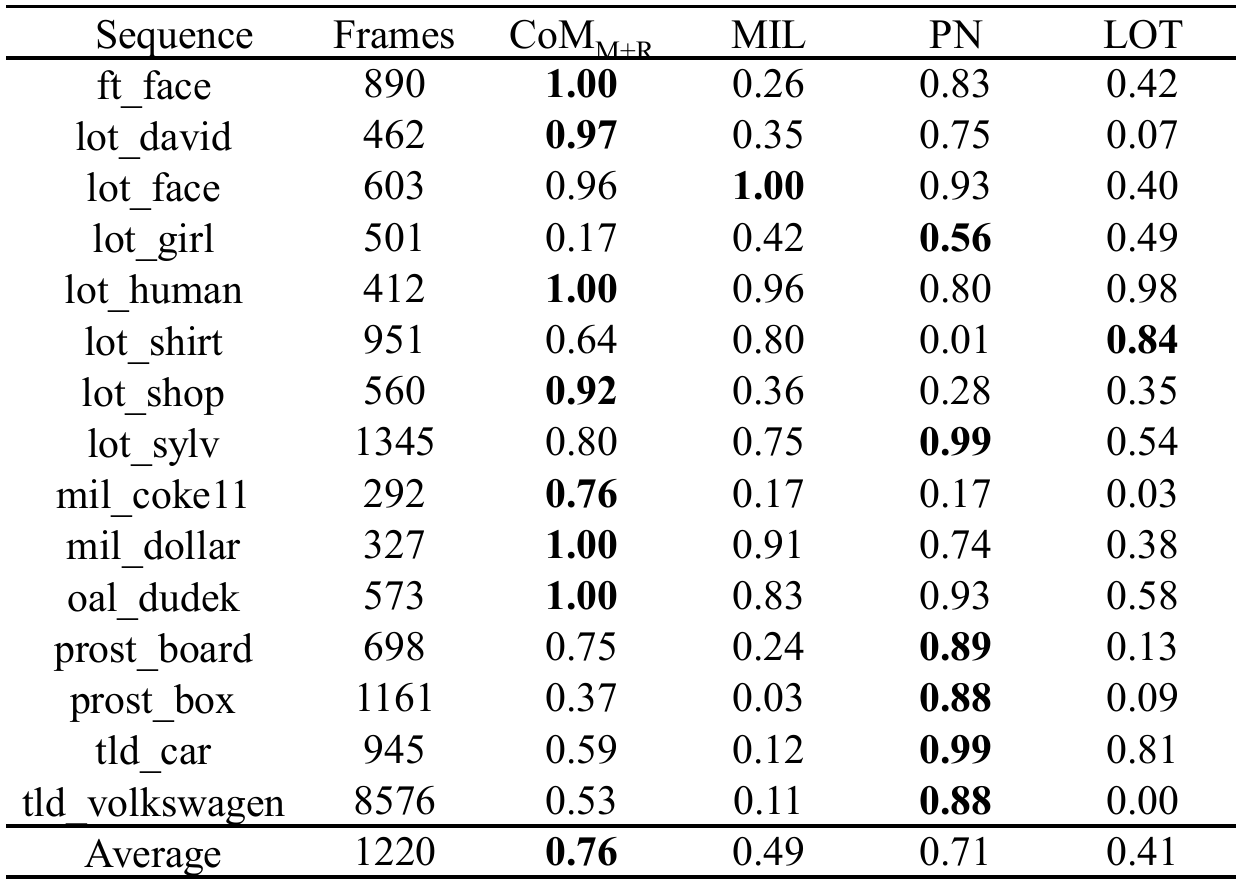}}
\end{center}
\caption{Comparison of successful tracking rate.}
\label{table:success_comp}
\end{table}

In this experiment we demonstrate the performance of our concurrent tracker ($CoT_{M+R}$) by comparing it with several recent state of the art trackers including: MIL tracker (MIL) \cite{Babenko09}, pn tracker (PN) \cite{Kalal10b}, and locally orderless tracker (LOT) \cite{Oron12}. In case of MIL tracker, scale of the detected bounding boxes are adjusted to be the same with the scale of ground truth for fair comparison as MIL tracker does not adopt to scale change. All code come from the original authors.

The quantitative comparisons are shown in Table \ref{table:accuracy_comp} and Table \ref{table:success_comp}. The best performance are shown in {\bf bold}. The results show that our concurrent tracker overall shows significant improvement of tracking performance especially in terms of tracking accuracy compared with state-of-the-art algorithms. It also is worth to note that the performance improvement is most significant  for the video sequences that involve partial or severe occlusions like 'ft\_face', 'lot\_human', 'lot\_shop', 'mil\_coke11', 'mil\_dollar', and 'oal\_dudek', demonstrating robustness against partial occlusions of the proposed tracker. In cases of 'prost\_board', 'prost\_box’, ’tld\_car’, ’tld\_volkswagen’ sequences for which pn tracker \cite{Kalal10b} obtains best score, the sequences contain challenging intervals where target objects disappear completely from camera view. Different from pn tracker, current implementation of the proposed tracker does not have detection functionality and failed to recover from such full disappearances.\\

\section{Conclusion}
\label{sec:conclusion}

In this paper we proposed a new tracking algorithm based on concurrent tracking of inliers and outliers. By tracking inliers and outliers concurrently, we are able to minimize possible misclassification errors regarding target state (occlusion/disocclusion), giving more reliable tracking performance. The experimental results shows that our tracker can track targets reliably without drift in various challenging scenarios, confirming the effectiveness of the proposed tracking paradigm.

One limit of our tracker is that it is hard to be recovered from the tracking failures due to moving target during full occlusion. Our future work is to extend the proposed tracker to cope with such failures by combining it with other detectors.



\begin{thebibliography}{99}

\bibitem{Kalal10a}
Kalal, Z., Mikolajczyk, K., Matas, J.: Forward-backward error: Automatic de-tection of tracking failures. In: International Conference on Pattern Recognition. (2010)

\bibitem{Adam06}
Adam, A., Rivlin, E., Shimshoni, I.: Robust fragments-based tracking using the integral histogram. In: Proc. IEEE Conference of Computer Vision and Pattern Recognition. (2006)

\bibitem{Shu12}
Shu, G., Dehghan, A., Oreifej, O., Hand, E., Shah, M.: Part-based multiple-person tracking with partial occlusion handling. In: Proc. IEEE Conference of Computer Vision and Pattern Recognition. (2012)

\bibitem{Hariharakrishnan05}
Hariharakrishnan, K., Schonfeld, D.: Fast object tracking using adaptive block matching. IEEE Trans. on Multimedia7(2005) 853-859

\bibitem{Pan07}
Pan, J., Hu, B.: Robust occlusion handling in object tracking. In: Proc. IEEE Conference of Computer Vision and Pattern Recognition. (2007)

\bibitem{Amezquita08}
Amezquita, N., Alquezar, R., Serratosa, F.: Dealing with occlusion in a probabilis-tic object tracking method. In: Proc. IEEE Conference of Computer Vision and Pattern Recognition. (2008)

\bibitem{Bouguet99}
Bouguet, J.Y.: Pyramidal implementation of the lucas kanade feature tracker: De-scription of the algorithm (1999) Technical Report, Intel Microprocessor Research Labs.

\bibitem{Fischler81}
Fischler, M., Bolles, R.: Random sample consensus: A paradigm for model tting with application to image analysis and automated cartography. Comm. Assoc. Comp24(1981) 381-395

\bibitem{Zhang12}
Zhang, C., Xu, J., Beaugendre, A., Goto, S.: A klt-based approach for occlusion handling in human tracking. In: Picture Coding Symposium (PCS). (2012)

\bibitem{Oron12}
Oron, S., Bar-Hillel, A., Levi, D., Avidan, S.: Locally orderless tracking. In: Proc. IEEE Conference of Computer Vision and Pattern Recognition. (2012)

\bibitem{Babenko09}
Babenko, B., Yang, M.H., Belongie, S.: Visual tracking with online multiple in-stance learning. In: Proc. IEEE Conf. Computer Vision and Pattern Recognition. (2009)

\bibitem{Ross08}
Ross, D.A., Lim, J., Lin, R.S., Yang, M.H.: Incremental learning for robust visual tracking. International Journal of Computer Vision (77). (2008)

\bibitem{Santner10}
Santner, J., Leistner, C., Saffari, A., Pock, T., Bischof, H.: Prost: Parallel ro-bust online simple tracking. In: Proc. IEEE Conf. Computer Vision and Pattern Recognition. (2010)

\bibitem{Kalal10b}
Kalal, Z., Matas, J., Mikolajczyk, K.: P-N learning: Bootstrapping binary classiers by structural constraints. In: Proc. IEEE Conf. Computer Vision and Pattern Recognition. (2010)

\bibitem{Everingham09}
Everingham, M., Gool, L.V., Williams, C.K.I., Winn, J., Zisserman, A.: The pascal visual object classes (voc) challenge. Int. J. Comput. Vision88(2009) 303-308

\end{thebibliography}
\end{document}